%% file: GG_Nav_fairmeta.tex
\PassOptionsToPackage{table}{xcolor}
\documentclass[]{fairmeta}

\usepackage[utf8]{inputenc}
\usepackage{amsmath}
\usepackage{amssymb}
\usepackage{amsfonts}
\usepackage{nicefrac}
\usepackage{makecell}
\usepackage{pifont}
\usepackage{array}
\usepackage{url}
\usepackage{eso-pic}

\newcommand{\cmark}{\ding{51}}
\newcommand{\xmark}{\ding{55}}

\definecolor{ggred}{RGB}{249,40,28}
\definecolor{navyellow}{RGB}{249,40,28}
\DeclareRobustCommand{\GGNav}{\texorpdfstring{\textcolor{ggred}{GG}\textcolor{navyellow}{-Nav}}{GG-Nav}}

\title{\GGNav{}: Goal-oriented Grounding Chain-of-Thought Elicits Reasoning in Navigation Foundation Models}

\author[1,2]{Shaohao Zhu}
\author[2,3]{Wensi Huang}
\author[2,4]{Siqi Zhang}
\author[2]{Yuqiang Yang}
\author[2,5]{Meng Wei}
\author[1,2]{Chenyang Wan}
\author[1]{Tianle Liu}
\author[2,*,\dagger]{Tai Wang}
\author[1,*]{Jinming Xu}

\affiliation[1]{Zhejiang University}
\affiliation[2]{Shanghai AI Laboratory}
\affiliation[3]{University of Science and Technology of China}
\affiliation[4]{Tongji University}
\affiliation[5]{The University of Hong Kong}

\contribution[*]{Corresponding author}
\contribution[\dagger]{Project lead}
\metadata[Project page]{\url{https://duangzhu.github.io/GG-Nav-Project-Page/}}
\metadata[Code]{\url{https://github.com/DuangZhu/GG_Nav}}

\abstract{%
Emerging strong Vision-Language Models (VLMs) have advanced navigation foundation models. However, these models still have unsatisfactory performance on goal-oriented navigation settings. In this paper, we first conduct a motivating study revealing that the key issue lies in the insufficient utilization of streaming observations and imprecise target localization. To tackle these challenges, we propose \textbf{\GGNav{}}, a simple yet effective navigation framework that learns a structured chain-of-thought (CoT) mechanism via supervised fine-tuning to explicitly localize goal-relevant instances, accumulate grounding evidence, and support more reliable decision-making. Specifically, the structured CoT is designed with two key stages: (i) Observation-to-Evidence, which reasons over all intermediate frames between adjacent decision steps to retain useful goal-related evidence; and (ii) Evidence-to-Decision, which integrates the current observation with accumulated evidence to ground the target and support the generation of navigation actions or clarification questions. Extensive experiments show that GG-Nav outperforms the state-of-the-art methods in terms of success rates on the HM3D ObjectNav, VL-LN, and CoIN benchmarks by 3.1\%, 16.6\%, and 5.3\%, respectively. Ablation studies further show that accumulating evidence from intermediate observations reduces evidence-missing failures by 11.6\%, while explicit grounding reduces stopping errors by 49.4\%. We also deploy GG-Nav on a Unitree Go2 robot, demonstrating its zero-shot generalization capability in unseen real-world environments.
}

\begin{document}

\AddToShipoutPictureFG*{%
  \begin{tikzpicture}[remember picture,overlay]
    \node[anchor=south west,xshift=3.0cm,yshift=-2.48cm] at (current page.north west) {\includegraphics[height=1.3cm]{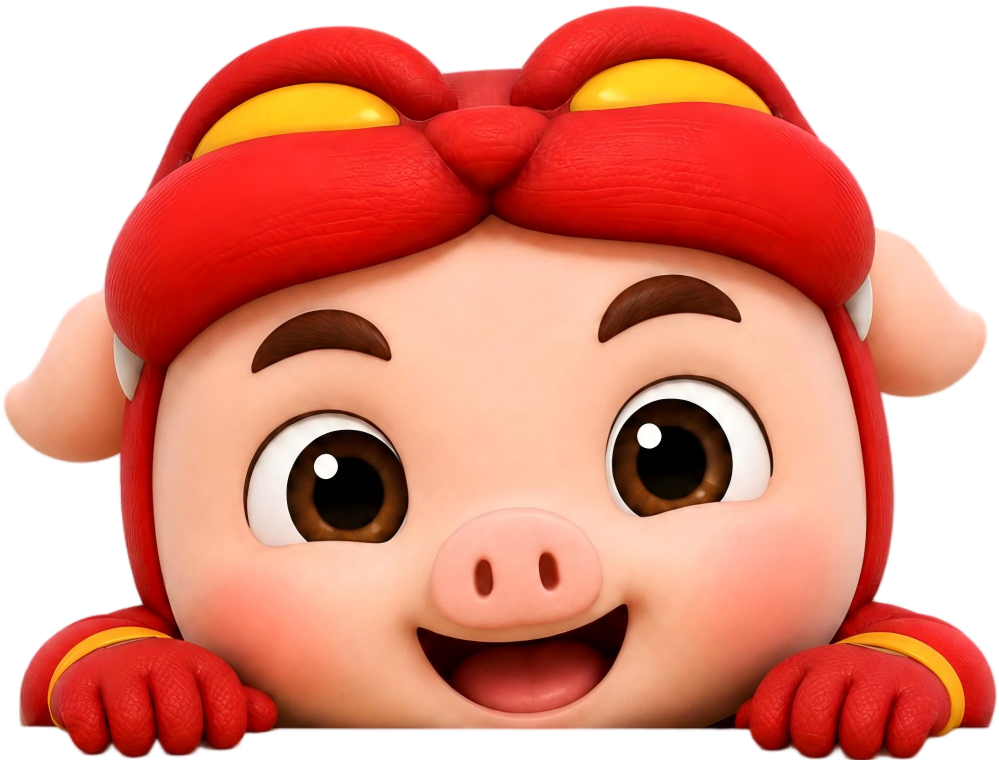}};
  \end{tikzpicture}%
}
\maketitle
\input{sections/Introduction}
\input{sections/Related_work}

\input{sections/Motivating_study}
\input{sections/Method}

\input{sections/experiment}

\input{sections/conclusion}

\bibliographystyle{plainnat}
\bibliography{references}

\clearpage
\appendix
\input{sections/appendix}

\end{document}

%% file: sections/Introduction.tex
\section{Introduction}

Goal-oriented navigation is a fundamental capability for embodied agents operating in human environments~\cite{yin2024sgnav,ieong2025survey}. Given a simple natural-language instruction, such as \emph{search for (the) chair}, a robot must explore an unseen area, interpret continuous egocentric observations, and make sequential decisions to reach the intended target. This problem includes object-goal navigation (ObjectNav)~\cite{batra2020objectnav,chaplot2020objectgoal,yu2026c,wang2026hydra}, where reaching any instance of the target category is sufficient, and interactive instance-goal navigation (IIGN)~\cite{CoIN,vlln}, where the agent must identify the intended target instance from an underspecified instruction by exploring the environment and interacting with humans. Compared with ObjectNav, IIGN is more practical but imposes stronger instance-level grounding requirements~\cite{vlln}: the agent must not only detect plausible candidate objects, but also identify the intended instance and maintain its identity consistently throughout navigation.

Recent advances in vision-language models (VLMs) have improved embodied navigation~\cite{wang2026hydra,zhang2024uni,zhang2025embodied}. Despite this progress, our detailed motivating study on goal-oriented navigation in Sec.~\ref{sec:motivating_study} reveals that current VLM-based methods still suffer from two major bottlenecks: they often miss target-related evidence in streaming observations and lack explicit target localization for precise stopping. To reduce decision latency, existing VLM-based methods typically select only sparse key observations for decision-making, rather than processing the full stream of visual frames~\cite{zhang2024uni, wei2025ground,yokoyama2025film,wei2025streamvln}. However, this sparse sampling strategy may overlook transient but crucial evidence of the target instance or distractors. Meanwhile, most VLM-based navigation methods directly map observations and language instructions to actions~\cite{zhang2024uni,zhang2025embodied,xiang2025nav}, leaving target grounding and localization implicit. As a result, the agent lacks an explicit estimate of where the target is, which makes it difficult to decide when to stop near the correct instance~\cite{batra2020objectnav,wang2026hydra,zhu2025move}. Although textual chain-of-thought reasoning has been explored~\cite{wang2026hydra,liu2025nav}, free-form rationales are often weakly grounded in visual instances, limiting their ability to preserve instance-level evidence and support fine-grained supervision~\cite{liu2025nav,sun2024emmax}.

\begin{figure*}[t]
    \centering
    \includegraphics[width=0.98\textwidth]{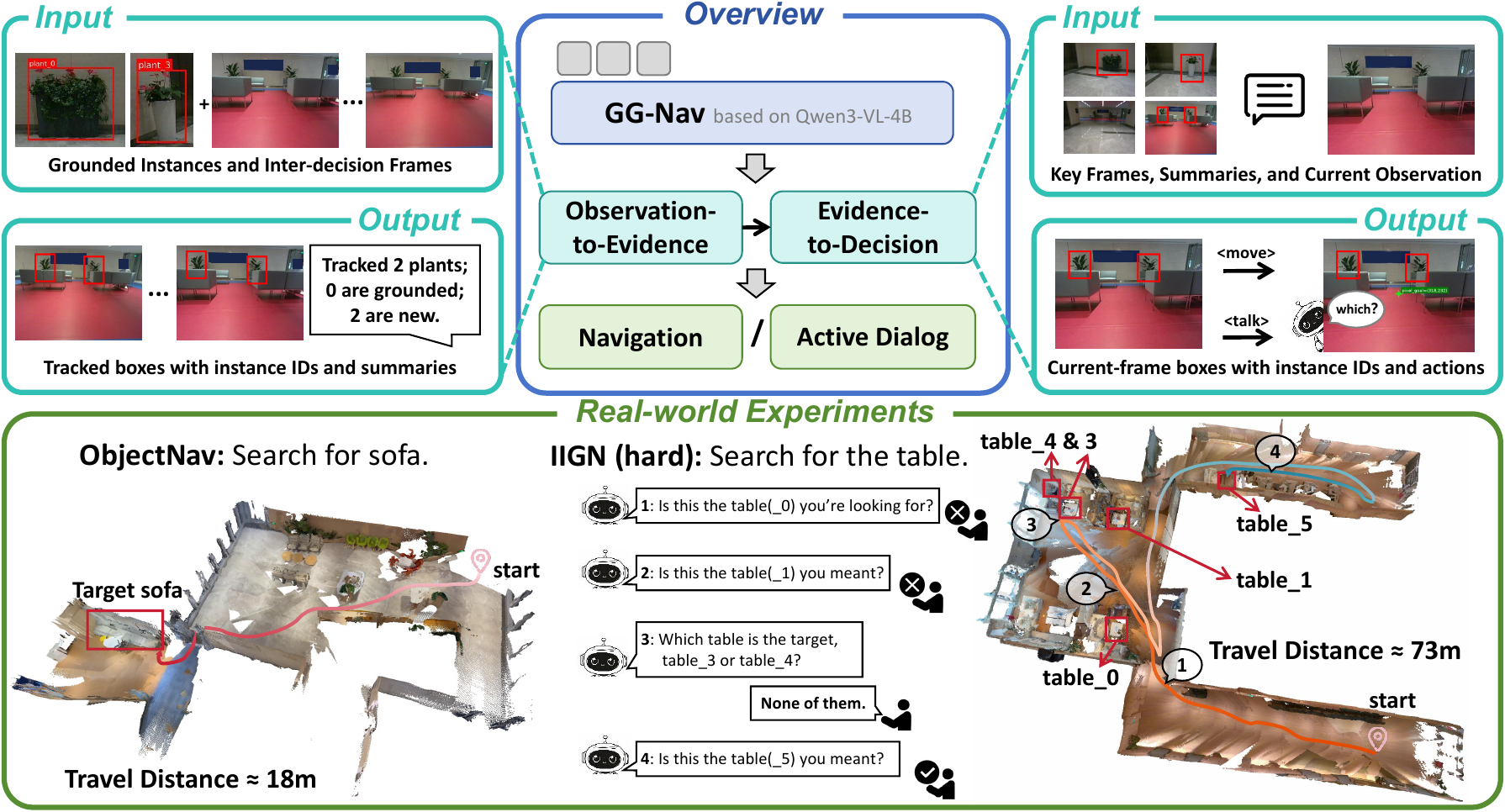}
    \caption{
        \textbf{Overview of GG-Nav.} Observation-to-Evidence converts intermediate frames between adjacent decision steps into accumulated grounding evidence. Evidence-to-Decision integrates the current observation with this evidence to ground goal-relevant instances and generate navigation actions or clarification questions. GG-Nav supports both ObjectNav and IIGN and supports real-world robot deployment.
    }
    \label{fig:teaser}
\end{figure*}

To address these issues, we propose \textbf{\GGNav{}},
a simple yet effective navigation framework that learns a \textbf{G}oal-oriented \textbf{G}rounding CoT mechanism via supervised fine-tuning to explicitly localize goal-relevant instances, accumulate grounding evidence, and support more reliable decision-making. 
As shown in Fig.~\ref{fig:teaser}, GG-Nav structures this process into two stages: Observation-to-Evidence and Evidence-to-Decision.

In each reasoning process, the Observation-to-Evidence stage first requires the model to accumulate key observations, retain useful key frames, ground goal-related evidence, and generate a summary over all intermediate frames collected between adjacent decision steps. Subsequently, the Evidence-to-Decision stage integrates the current observation with the accumulated evidence to explicitly ground goal-relevant instances and generate either a navigation action or a clarification question. To learn this structured reasoning mechanism, we construct a fine-tuning dataset with 30K grounding-annotated videos for evidence extraction and instance tracking, together with 60K navigation trajectories for evidence-conditioned decision-making to fine-tune the foundation VLM. In this way, GG-Nav is equipped with the capability of producing instance bounding boxes and consistent instance IDs, making its intermediate reasoning procedure spatially grounded and visually verifiable.

We evaluate GG-Nav on HM3D~\cite{ramakrishnan2021hm3d} for ObjectNav and on VL-LN~\cite{vlln} and CoIN~\cite{CoIN} for IIGN. GG-Nav achieves the state-of-the-art performance across all three benchmarks, outperforming the strongest competing methods by 3.1\%, 16.6\%, and 5.3\%, respectively. Progressive ablations show that incorporating intermediate observations into decision-making improves model performance by reducing failures caused by missing target-related evidence by 11.6\%. In addition, explicitly generating bounding-box grounding results improves target localization and enables the agent to stop at more precise positions, reducing stopping errors by 49.4\%. The structured grounding outputs further support stage-wise failure analysis, revealing that exploration and instance grounding remain the main bottlenecks. Finally, as shown in Fig.~\ref{fig:teaser}, we deploy GG-Nav on a Unitree Go2 robot and demonstrate its zero-shot capability in unseen real-world environments under both ObjectNav and IIGN settings.

%% file: sections/Related_work.tex
\section{Related Work}
\subsection{Goal-oriented Navigation Benchmarks}
Goal-oriented navigation requires an embodied agent to reach a user-specified target, serving as a foundational skill for real-world tasks such as retrieval, delivery, and inspection. This paper focuses on text-instructed goal-oriented navigation, where the target is specified in natural language. Early benchmarks mainly focused on object-goal navigation (ObjectNav), where the agent is instructed to find any instance of a target category, such as ``find a chair'' \cite{batra2020objectnav}. While this setting provides a useful benchmark for goal-oriented navigation, it does not fully cover everyday robot-assistance scenarios, where users often refer to a specific or personally relevant instance. More recent work further considers instance-goal navigation, where the target is a specific object instance rather than any category member \cite{khanna2024goatbench}. However, these settings often assume that users can provide detailed and unambiguous target descriptions, which may not hold in daily human-robot interaction. To better match realistic human-robot interaction, interactive instance-goal navigation (IIGN) was introduced, where users may provide underspecified instructions (Search for the chair) and the agent must resolve the target ambiguity through interaction with a human during exploration \cite{vlln,CoIN}. From this perspective, ObjectNav and IIGN represent two practical forms of goal-oriented navigation for everyday robot assistance. Despite their different goal specifications, both require the agent to accumulate goal-relevant evidence from continuous observations and ground it to candidate visual targets for reliable decision-making. This motivates our unified framework, which supports both settings through explicit evidence accumulation and grounding.

\subsection{Goal-oriented Navigation Methods}

\textbf{Classical methods.}
Before the emergence of large foundation models, goal-oriented navigation methods mainly followed two paradigms: end-to-end and modular approaches. End-to-end methods directly map instructions and observations to navigation actions through reinforcement learning, imitation learning, or hybrid training strategies~\cite{ye2021auxiliary,ramrakhya2022habitatweb,maksymets2021thda,ramrakhya2023pirlnav}. Progress in ObjectNav was driven by auxiliary objectives~\cite{ye2021auxiliary}, large-scale human demonstrations~\cite{ramrakhya2022habitatweb}, data augmentation~\cite{maksymets2021thda}, and BC-to-RL training~\cite{ramrakhya2023pirlnav}. In parallel, modular methods decomposed navigation into semantic mapping, high-level goal selection, and local planning, enabling agents to navigate via structured semantic maps~\cite{ramakrishnan2022poni,georgakis2021learning}.

\textbf{Foundation-model-based methods.} With the rise of VLMs and LLMs, recent work has incorporated foundation models into goal-oriented navigation along three main lines. 
The first line follows a modular paradigm, using foundation models as components for commonsense reasoning~\cite{long2024instructnav,kuang2024openfmnav,yin2025unigoal}, open-vocabulary perception~\cite{yokoyama2024vlfm,zhang2024trihelper,zhou2025beliefmapnav}, scene or memory summarization~\cite{yin2024sgnav,wu2024voronav,cao2025cognav,nie2025wmnav}, and high-level waypoint selection~\cite{yin2024sgnav,yu2023l3mvn,yuan2024gamap,cai2024bridging}. 
Although effective, these methods typically use foundation models in a zero-shot manner, resulting in weak alignment with embodied navigation, and their modular pipelines are prone to error accumulation in complex scenes~\cite{zhang2025embodied,yokoyama2024vlfm,zhang2024trihelper,nie2025wmnav}.

The second line directly adapts foundation models to goal-oriented navigation via supervised fine-tuning on embodied trajectories, learning end-to-end policies from observations and instructions to actions or exploration decisions. Representative methods include Uni-NaVid~\cite{zhang2024uni}, Hydra-Nav~\cite{wang2026hydra}, NavFoM~\cite{zhang2025embodied}, and Nav-$R^2$~\cite{xiang2025nav}. 
These methods better leverage semantic priors and reduce error accumulation from modular pipelines, but placing large VLMs in the control loop tightly couples high-level reasoning with low-level control, limiting real-time and smooth long-horizon navigation~\cite{wei2025ground}. 
To mitigate this issue, a third line adopts dual-system designs, where VLMs perform sparse high-level reasoning while lightweight controllers handle fast local execution, as in Nav-R1~\cite{liu2025nav}, FiLM-Nav~\cite{yokoyama2025film}, and InternVLA-N1~\cite{wei2025ground}.

Despite these advances, existing dual-system methods still face two key challenges. First, sparse VLM reasoning improves efficiency but leaves many intermediate observations unused, even though they may contain transient evidence of the target or distractors. Second, target localization and spatial reasoning often remain implicit in the decision-making process. In this paper, we address these limitations by learning a structured goal-oriented CoT mechanism.

%% file: sections/Motivating_study.tex
\section{Motivating Study}
\label{sec:motivating_study}

In this section, we aim to conduct a quantitative analysis on top of a basic adapted navigation foundation model to investigate the key challenges in goal-oriented navigation.

Existing methods typically construct visual history by uniformly sampling a small number of past frames and adding dialogue-related key frames. For example, InternVLA-N1~\cite{internvla-n1} uses eight uniformly sampled history frames together with interaction frames as visual context. We fine-tune the System~2 component of InternVLA-N1 on a targeted goal-oriented navigation dataset and evaluate it on 500 episodes from VL-LN Bench~\cite{vlln}. This baseline achieves a success rate of 25.4\%.

\begin{table}[htbp!]
    \centering
    \caption{\textbf{Failure analysis of adapted InternVLA-N1 on VL-LN Bench.}}
    \label{tab:motivating_failure_analysis}
    \setlength{\tabcolsep}{4pt}
    \renewcommand{\arraystretch}{1.12}
    \small
    \begin{tabular}{llcc}
        \toprule
        Target visibility & Error type & \# Failures & Ratio \\
        \midrule
        \makecell[l]{Not observed\\119 (32\%)} 
            & Exploration error & 119 & 32\% \\
        \midrule
        \multirow{3}{*}{\makecell[l]{Target appears\\ in observations\\254 (68\%)}}
            & Evidence miss   & 153 & 41\% \\
            & Reaching error  & 26  & 7\%  \\
            & Stopping error  & 75  & 20\% \\
        \midrule
        \multicolumn{2}{l}{Total failures}
            & 373 & 100\% \\
        \bottomrule
    \end{tabular}
\end{table}

As summarized in Tab.~\ref{tab:motivating_failure_analysis}, we further analyze the 373 failure cases and find that the target object actually appears in the agent's RGB observations in 68\% (254) of failed episodes. Among all failures, 41\% (153) occur because the target appears only briefly in intermediate frames that are not selected by the sparse history sampling strategy, causing the model to miss critical evidence. Another 20\% (75) correspond to cases where the model detects and even confirms the target through interaction but still fails to stop accurately near the target. This indicates that implicit action prediction is not sufficient to translate detected evidence into precise target localization and reliable navigation. These observations suggest that applying navigation foundation models to goal-oriented navigation still faces two key challenges: discovering goal-relevant evidence from streaming observations, and grounding and reusing accumulated evidence to localize the target and support reliable decision-making. These two challenges motivate the method presented next.

%% file: sections/Method.tex
\section{The Proposed Framework}
\begin{figure*}[t]
    \centering
    \includegraphics[width=\textwidth]{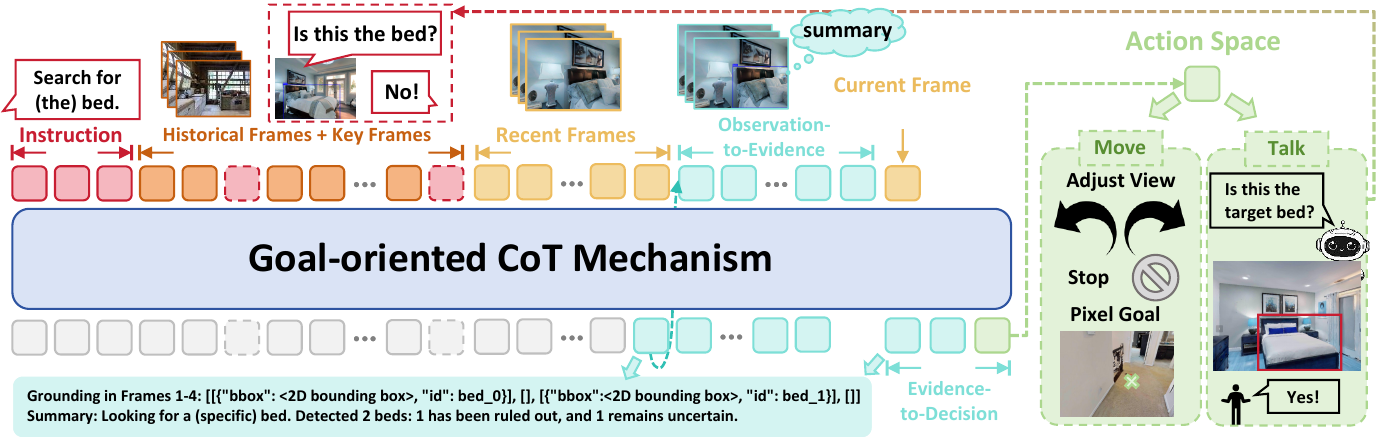}
    \caption{\textbf{Inference process of \GGNav{}.} Given the instruction, compact visual history, dialogue-associated evidence, and intermediate frames since the previous decision step, GG-Nav first converts the intermediate frames into accumulated grounding evidence through Observation-to-Evidence, and then uses Evidence-to-Decision to integrate this evidence with the current observation and generate either a navigation action or a clarification question.}
    \label{fig:framework_overview}
\end{figure*}
\subsection{Task Formulation}

We consider two text-instructed goal-oriented navigation tasks: ObjectNav~\cite{batra2020objectnav} and IIGN~\cite{vlln}. Given an instruction $I_c$ specifying a target category $c$, let $\mathcal{G}(I_c)$ denote the set of valid goal instances. In ObjectNav, any instance of category $c$ is a valid goal, whereas IIGN requires the agent to identify a unique target instance $v^\ast$ among multiple same-category candidates. The agent must therefore accumulate visual evidence during exploration and may ask a human oracle questions to disambiguate the target instance.

As shown in Fig.~\ref{fig:framework_overview}, the action space consists of navigation actions $\mathcal{A}^{\mathrm{move}}$, including view adjustment, pixel-goal navigation, and \texttt{STOP}. In IIGN, the agent may additionally choose a dialogue action from $\mathcal{A}^{\mathrm{talk}}$, represented as a natural-language question grounded in one or more observed candidate instances. The oracle's answer is then appended to the dialogue history for subsequent decision-making. An episode is considered successful if the agent issues \texttt{STOP} within the allowed step budget and terminates sufficiently close to a valid goal instance in $\mathcal{G}(I_c)$. As a result, ObjectNav regards any instance of category $c$ as a valid goal, whereas IIGN accepts only the intended target instance $v^\ast$.

\subsection{Goal-oriented Grounding CoT Mechanism}
\noindent\textbf{Goal-oriented CoT Formulation.} Building on the above analysis, GG-Nav aims to equip a VLM with a structured reasoning mechanism that discovers goal-relevant evidence from streaming observations and grounds accumulated evidence for reliable decision-making. For implementation, we build GG-Nav upon Qwen3-VL-4B~\cite{bai2025qwen3} and formulate each decision step $i$ as a structured reasoning trace $\mathcal{C}_i^{\mathrm{goal}}=(R_i^{\mathrm{OE}},R_i^{\mathrm{ED}},a_i)$, which decomposes the reasoning process into Observation-to-Evidence~(OE) and Evidence-to-Decision~(ED). Specifically, $R_i^{\mathrm{OE}}$ revisits intermediate observations to recover goal-relevant evidence through frame-wise grounding records and evidence summaries; $R_i^{\mathrm{ED}}$ grounds the current observation and associates visible candidates with accumulated evidence; and $a_i$ denotes the resulting action.

\subsubsection{\textbf{Observation-to-Evidence}}
\label{sec:instance_tracking_stage}
Observation-to-Evidence is triggered when the agent reaches a new decision step and invokes the VLM. Its inputs include the instruction, a compact set of historical frames, key frames with grounded instances, and intermediate frames collected since the previous decision step. The compact set of historical frames consists of frames uniformly sampled between the beginning of the task and the previous decision step. Each key frame is paired with its grounding result and, if available, its question-answering record, informing the model of previously examined candidates and providing visual evidence for instance tracking. This context enables the model to associate newly observed candidates with recorded instances.

Conditioned on this context, the model reasons over every intermediate frame and generates an Observation-to-Evidence CoT ($R_i^{\mathrm{OE}}$). For each frame, it produces a structured JSON record containing the 2D bounding boxes and temporally consistent IDs of all instances belonging to the instructed category. It then summarizes which instances have been previously encountered and which are newly observed. These frame-wise grounding records and summaries transform transient inter-decision observations into explicit evidence, allowing the model to better exploit the streaming observation.

\subsubsection{\textbf{Evidence-to-Decision}}
\label{sec:decision_making_stage}
This stage integrates the accumulated grounding evidence with the current observation to support action generation. The model first produces a current-frame grounding result, explicitly predicting the bounding boxes and consistent IDs of goal-relevant instances and associating them with the accumulated evidence $R_i^{\mathrm{ED}}$. It then conditions on all the information to generate the next decision ($a_i$). In ObjectNav, the output is restricted to $a_i\in\mathcal{A}^{\mathrm{move}}$; in IIGN, $a_i\in\mathcal{A}^{\mathrm{move}}\cup\mathcal{A}^{\mathrm{talk}}$. For dialogue actions, the model asks a question grounded in selected intermediate or current frames and their instance IDs.
After receiving a reply from the human or oracle, the question-answer pair is stored with the grounded frame as evidence for accumulation and subsequent decision making. This stage provides a spatially verifiable basis for target localization and decision-making.

\subsection{Data Collection}
\label{sec:data_collection}
We fine-tune GG-Nav with two sources of task-specific supervision, namely \emph{videos with grounding annotations} and \emph{navigation trajectory data}, to better follow the proposed goal-oriented CoT mechanism. The former provides dense supervision for Observation-to-Evidence by supervising instance grounding and tracking without requiring action annotations, while the latter supervises the full task execution process.

\textbf{Videos with grounding annotations.} To enhance temporal grounding and evidence accumulation, we construct training samples from video streams with object-level grounding annotations.
We use the Aria Synthetic Environments Dataset, which provides continuous egocentric videos with object segmentation masks.
For each video, we first sample an object category $c$ that appears in the scene. We then sample $K$ context frames containing at least one instance of category $c$ and extract a contiguous clip of $L$ frames from the same video stream.
Each context frame is paired with the grounding annotations of the selected category, including 2D bounding boxes and instance IDs.
An Observation-to-Evidence sample can be represented as
\[ s^{\mathrm{oe}} := \left\{ \left\{ I_c,\ \{(\bar{o}_j,\bar{g}_j)\}_{j=1}^{K},\ \{o_t\}_{t=1}^{L} \right\} \rightarrow \left\{ \{g_t\}_{t=1}^{L},\ r \right\} \right\}, \]
where the left side denotes the input, with $I_c$ representing the category-level instruction, $(\bar{o}_j,\bar{g}_j)$ the $j$-th context frame and its grounding annotation, and $o_t$ the $t$-th intermediate frame, while the right side denotes the output, with $g_t$ representing the corresponding structured grounding record and $r$ the accumulated-evidence summary.
This construction follows the format of Observation-to-Evidence and enables the use of video data without action labels.

\textbf{Navigation trajectory data.} To jointly supervise all stages in GG-Nav, we further collect ObjectNav and IIGN trajectories in Habitat~\cite{savva2019habitat}. Following prior work~\cite{vlln}, we use annotated scenes from the training splits of HM3D~\cite{ramakrishnan2021hm3d} and MP3D~\cite{chang2017matterport3d}. For each episode, we use a frontier-based exploration algorithm to iteratively select frontier waypoints as navigation targets. The selected waypoints are connected by an A$^\ast$ planner~\cite{hart1968formal} to obtain collision-free paths, which are then converted into executable low-level actions.

During data collection, we record RGB observations, agent poses, action labels, and the 2D bounding boxes and instance IDs of all visible instances belonging to the target category. 
For IIGN episodes, we construct interaction supervision by generating question-answer pairs with a rule-based procedure. The question is triggered only when the agent observes a previously unseen candidate instance from the target category. The answer is then produced by an automated oracle according to the ground-truth target identity. 
The resulting question-answer pair is added to the data as dialogue history.

After generating trajectories, we split each trajectory into decision-step segments and use each segment as a training sample. Formally, the $i$-th training segment is represented as
\[
s_i^{\mathrm{nav}}
:=
\left\{
\left\{
I_c,\ 
H_i,\ 
D_i,\ 
O_i^{\mathrm{rec}},\ 
o_i^{\mathrm{cur}}
\right\}
\rightarrow
\left\{
R_i^{\mathrm{OE}},\ 
R_i^{\mathrm{ED}},\ 
a_i
\right\}
\right\},
\]
where $H_i$ denotes the compact historical visual memory, $D_i$ denotes the accumulated dialogue history, $O_i^{\mathrm{rec}}$ is the sequence of intermediate frames retained since the previous decision, and $o_i^{\mathrm{cur}}$ is the current observation. $R_i^{\mathrm{OE}}$ denotes the Observation-to-Evidence output, $R_i^{\mathrm{ED}}$ denotes the current grounding component of Evidence-to-Decision, and $a_i$ is the supervised navigation or interaction action.

In particular, in this work, we generate approximately 30K videos with grounding annotations and 60K navigation trajectories, comprising 20K ObjectNav trajectories and 40K IIGN trajectories, to fine-tune the model.

%% file: sections/experiment.tex
\section{Experiments}

Our experiments are designed to answer four questions:
1) whether GG-Nav improves goal-oriented navigation;
2) whether intermediate observations reduce failures caused by missed evidence;
3) whether explicit grounding improves target localization and stopping; and
4) whether the learned mechanism transfers to unseen real-world environments.

\textbf{Evaluation metrics.}
We evaluate both ObjectNav and IIGN using standard navigation metrics.
\textbf{SR} measures whether the agent successfully stops within $0.25$\,m of a target object.
\textbf{SPL} additionally accounts for path efficiency by weighting successful episodes according to the ratio of the shortest path length to the executed path length.
For IIGN, we additionally report \textbf{NQ}, the average number of questions asked per episode, to quantify the interaction cost of instance disambiguation.

\textbf{Implementation details.}
We initialize GG-Nav from Qwen3-VL-4B-Instruct and fully fine-tune the vision encoder, multimodal projector, and language-model backbone for 2.5 epochs. We use AdamW~\cite{loshchilov2019decoupled} with a learning rate of $2 \times 10^{-5}$ for the language-model backbone and multimodal projector, a smaller learning rate of $5 \times 10^{-6}$ for the vision encoder, and a global batch size of 128. Additional training details are provided in Appendix~\ref{app:implementation}.

\begin{table*}[htbp!]
\centering
\small
\setlength{\tabcolsep}{4pt}
\renewcommand{\arraystretch}{1.15}
\caption{\textbf{Baseline comparison on HM3D ObjectNav and the VL-LN and CoIN IIGN benchmarks.}}
\label{tab:main_results}
\resizebox{\textwidth}{!}{
\begin{tabular}{l c !{\vrule width 0.6pt} cc !{\vrule width 0.6pt} ccc !{\vrule width 0.6pt} ccc !{\vrule width 0.6pt} ccc !{\vrule width 0.6pt} ccc}
\toprule
\multirow{2}{*}{Method} & \multirow{2}{*}{Interaction}
& \multicolumn{2}{c|}{HM3Dv2}
& \multicolumn{3}{c|}{VL-LN} 
& \multicolumn{3}{c|}{CoIN (Seen)} 
& \multicolumn{3}{c|}{CoIN (Seen Synonyms)} 
& \multicolumn{3}{c}{CoIN (Unseen)} \\
\cmidrule(lr){3-4}
\cmidrule(lr){5-7}
\cmidrule(lr){8-10}
\cmidrule(lr){11-13}
\cmidrule(lr){14-16}
& 
& SR $\uparrow$& SPL $\uparrow$
& SR $\uparrow$& SPL $\uparrow$& NQ $\downarrow$
& SR $\uparrow$& SPL $\uparrow$& NQ $\downarrow$
& SR $\uparrow$& SPL $\uparrow$& NQ $\downarrow$
& SR $\uparrow$& SPL $\uparrow$& NQ $\downarrow$ \\
\midrule

FBE       & \xmark & 31.5 & 13.69 & 9.2 & 5.09 & - & 4.69 & 2.75 & - & 2.35 & 1.55 & - & 3.92 & 2.05 & - \\
VLFM~\cite{yokoyama2024vlfm}      & \xmark & 63.6 & 32.5 & 8.8 & 5.63 & - & 0.48 & 0.29 & - & 0.28 & 0.02 & - & 0.22 & 0.10 & - \\
MTU3D~\cite{zhu2025move}     & \xmark & 59.3 & 19.87 & 12.8 & 5.44 & - & 27.79 & 11.23 & - & 26.54 & 12.95 & - & 8.71 & 3.34 & -  \\

\midrule

InternVLA-N1~\cite{internvla-n1}     & \cmark & 69.2 & 27.61 & 25.4 & 16.18 & 1.98 & 17.57 & 9.51 & 2.79 & 20.94 & 11.13 & 2.54 & 17.21 & 8.20 & 2.61 \\
AIUTA~\cite{CoIN}     & \cmark & - & - & 4.0 & 2.16 & 5.42 & 7.42 & 2.92 & 1.67 & 14.38 & 7.99 & 1.36 & 6.67 & 2.30 & 1.13 \\
GG-Nav      & \cmark & \textbf{72.3} & \textbf{31.12} & \textbf{42.0} & \textbf{23.22} & 3.01 & \textbf{29.96} & \textbf{13.54} & 4.66 & \textbf{31.86} & \textbf{16.62} & 4.92 & \textbf{19.61} & \textbf{8.82} & 1.79 \\

\bottomrule
\end{tabular}
}
\end{table*}

\subsection{Baseline Comparison}
We evaluate GG-Nav on three benchmarks: HM3D for ObjectNav, and VL-LN and CoIN for IIGN. We compare it with representative non-interactive ObjectNav baselines, including FBE, VLFM~\cite{yokoyama2024vlfm}, and MTU3D~\cite{zhu2025move}, as well as interactive IIGN baselines, including InternVLA-N1~\cite{internvla-n1} and AIUTA~\cite{CoIN}. Non-interactive methods only execute navigation actions, while interactive methods can additionally ask clarification questions.

As shown in Tab.~\ref{tab:main_results}, GG-Nav achieves the best overall performance under both ObjectNav and IIGN settings. Specifically, on the HM3D ObjectNav benchmark, GG-Nav obtains 72.3\% SR and 31.12\% SPL, improving over the strongest baseline, InternVLA-N1, by 3.1\% in SR and 3.51\% in SPL. On the VL-LN benchmark, GG-Nav achieves 42.0\% SR, outperforming the second-best method by 16.6\%. Moreover, on all three evaluation splits of CoIN, GG-Nav consistently achieves the best SR and SPL. These consistent gains demonstrate that GG-Nav provides a unified and effective framework for both ObjectNav and IIGN.

Modular methods remain limited in solving the IIGN problem. Specifically, even with careful parameter tuning, AIUTA achieves only 4\%, 7.42\%, and 6.67\% on the VL-LN benchmark and the seen/unseen splits of the CoIN benchmark, respectively. This limitation is mainly because existing modular methods construct semantic maps based on image-level scores, which can indicate whether a view contains the target category but cannot explicitly localize or bind a specific instance. As a result, when multiple candidates appear in an image, these methods may fail to distinguish the correct target, leading to incorrect navigation decisions. In addition, introducing more specialized modules for questioning and object tracking can make the modular pipeline more fragile, because errors may accumulate across modules and eventually result in incorrect navigation decisions.

\subsection{Ablation Study I: Fundamental Designs}
We conduct a progressive ablation study to verify the contribution of each design component. The action-only baseline takes the instruction, historical key frames, and dialogue-associated frames as context and directly predicts the next action. We then add intermediate frames as additional observations, while still requiring the model to output only the final action. The third variant further introduces Observation-to-Evidence, which converts intermediate frames into structured grounding evidence, while the full GG-Nav model additionally performs Evidence-to-Decision to integrate this evidence for action generation.

\begin{table*}[htbp!]
    \centering
    \small
    \setlength{\tabcolsep}{5pt}
    \renewcommand{\arraystretch}{1.15}
    \caption{\textbf{Progressive ablation study.}}
    \label{tab:ablation_progressive}
    \resizebox{\textwidth}{!}{
    \begin{tabular}{lccc cc ccc ccc}
    \toprule
    \multirow{2}{*}{Variant}
    & \multicolumn{3}{c}{Setting}
    & \multicolumn{2}{c}{HM3D}
    & \multicolumn{3}{c}{VL-LN}
    & \multicolumn{3}{c}{CoIN (Seen Synonyms)} \\
    \cmidrule(lr){2-4}
    \cmidrule(lr){5-6}
    \cmidrule(lr){7-9}
    \cmidrule(lr){10-12}
    & Inter. Frames & O-to-E & E-to-D
    & SR $\uparrow$ & SPL $\uparrow$
    & SR $\uparrow$ & SPL $\uparrow$ & NQ $\downarrow$
    & SR $\uparrow$ & SPL $\uparrow$ & NQ $\downarrow$ \\
    \midrule
    
    Action-only
    & \xmark & \xmark & \xmark
    & 62.5 & 21.56
    & 20.8 & 10.56 & 2.44
    & 18.0 & 11.99 & 2.14 \\
    
    + Intermediate frames
    & \cmark & \xmark & \xmark
    & 66.7 & 25.78
    & 36.8 & 20.39 & 2.79
    & 28.3 & 18.13 & 3.82 \\
    
    + Observation-to-Evidence
    & \cmark & \cmark & \xmark
    & 68.9 & 26.94
    & 36.6 & 20.08 & 2.80
    & 27.7 & 14.68 & 3.66 \\
    
    + Evidence-to-Decision
    & \cmark & \cmark & \cmark
    & 72.3 & 31.12
    & 42.0 & 23.22 & 3.01
    & 31.9 & 16.62 & 4.92 \\
    
    \bottomrule
    \end{tabular}
    }
    \begin{minipage}{\textwidth}
    \footnotesize
    \textbf{Note.}
    `Inter. Frames' denotes using intermediate frames; O-to-E and E-to-D denote Observation-to-Evidence and Evidence-to-Decision.

    \end{minipage}
    \end{table*}

As shown in Tab.~\ref{tab:ablation_progressive}, our full model achieves the highest success rates across all three benchmarks, indicating the effectiveness of our overall design. Compared with the action-only baseline, the full model improves SR from 62.5\% to 72.3\% on HM3D, from 20.8\% to 42.0\% on VL-LN, and from 18.0\% to 31.9\% on CoIN. The gains in SPL are also substantial on HM3D and VL-LN, increasing from 21.56\% to 31.12\% and from 10.56\% to 23.22\%, respectively. These results show that intermediate observations provide useful evidence beyond key frames, while Observation-to-Evidence and explicit current-frame grounding within Evidence-to-Decision further improve decision making. We also observe that the full model asks more questions, suggesting a trade-off between navigation success and interaction efficiency in ambiguous scenes. 

To better understand failure modes, we conduct a fully automated stage-wise error analysis using simulator instance-segmentation observations, agent grounding/tracking logs, and final poses. Each failed episode is attributed to one pipeline stage: \emph{exploration}, \emph{grounding}, \emph{tracking}, \emph{reaching}, or \emph{stopping}. Exploration errors occur when the target instance never appears in the simulator's observations; grounding errors when it appears but is never associated with the target instance ID in the model's grounding output; tracking errors when a previously grounded target is later lost or switched to another instance; reaching errors when the correct target association is maintained but the final position is at least $1$\,m from the target; and stopping errors when the final position is within $1$\,m but more than $0.25$\,m from the target.

\begin{table}[htbp!]
\centering
\footnotesize
\setlength{\tabcolsep}{4pt}
\renewcommand{\arraystretch}{1.08}
\caption{\textbf{Stage-wise error analysis across benchmarks.}}
\label{tab:error_analysis_stagewise}
\begin{tabular}{llccccc}
\toprule
Benchmark & Variant & E & G & T & R & S \\
\midrule
\multirow{4}{*}{HM3D}
& Action-only          & 111  & 162  & - & - & 102 \\
& + Intermediate frames  & 107 & 141 & - & - & 85  \\
& + Observation-to-Evidence        & 93  & 154  & 6 & 20 & 38 \\
& + Evidence-to-Decision         & 92  & 122  & 4 & 27 & 32 \\
\midrule
\addlinespace[1mm]

\multirow{4}{*}{VL-LN}
& Action-only          & 163  & 164 & - & 54 & 15 \\
& + Intermediate frames  & 123 & 143 & - & 23 & 27 \\
& + Observation-to-Evidence         & 126  & 147 & 22 & 9 & 13 \\
& + Evidence-to-Decision         & 137  & 103 & 15 & 17 & 18 \\
\midrule
\addlinespace[1mm]

\multirow{4}{*}{CoIN}
& Action-only          & 32 & 155 & - & 51 & 40 \\
& + Intermediate frames & 26 & 141 & - & 17 & 59 \\
& + Observation-to-Evidence        & 21 & 153 & 39 & 13 & 19 \\
& + Evidence-to-Decision          & 29 & 126 & 23 & 25 & 28 \\
\bottomrule
\end{tabular}

\begin{minipage}{0.53\linewidth}
\footnotesize
\textbf{Note.}
Each failed episode is attributed to the stage in the pipeline:
Exploration (E), Grounding (G), Tracking (T), Reaching (R), and Stopping (S).
\end{minipage}
\end{table}

Introducing intermediate observations consistently reduces grounding errors by enabling the agent to exploit recently observed images for decision making. Specifically, compared with the `Action-only' variant, the `+ Intermediate frames' variant decreases grounding errors from 162 to 141 on HM3D, from 164 to 143 on VL-LN, and from 155 to 141 on CoIN, corresponding to an average reduction of 11.6\%. These reductions suggest that a non-negligible portion of grounding failures stems from the limited use of observations under sparse decision making: the target instance may appear in frames between two decision steps, but this evidence is unavailable to the action-only baseline when predicting the next action. By incorporating intermediate observations, the agent can carefully examine such skipped frames and better ground potential targets.

Explicit grounding also improves target localization and enables the agent to stop at more precise positions. This effect is reflected in the reduction of stopping errors when moving from the `+ Intermediate frames' variant to the full model: the number of stopping errors decreases from 85 to 32 on HM3D, from 27 to 18 on VL-LN, and from 59 to 28 on CoIN, corresponding to an average reduction of 49.4\%. These results suggest that explicit target grounding provides a concrete visual basis for both predicting \texttt{STOP} and selecting navigation waypoints, leading the agent to stop more accurately.

Despite these improvements, the remaining failures are still primarily concentrated in exploration and grounding errors. In the full model, these two categories account for 77.3\% of all failures on HM3D, 82.8\% on VL-LN, and 67.1\% on CoIN. This distribution indicates that the major bottlenecks lie in long-term memory and instance-level grounding. Further progress therefore likely requires stronger memory mechanisms that help the agent identify under-explored regions more effectively, as well as more robust grounding capability that can distinguish and maintain target instances throughout navigation.

\subsection{Ablation Study II: Dialog-frame Retention}
\label{sec:grounding_evidence_retention}

To support long-horizon instance grounding, GG-Nav accumulates grounding evidence over time and reuses it as memory for later instance tracking and decision making. However, continuously retaining all historical evidence leads to an ever-growing memory, which increases inference cost and may introduce noise into decision-making. Therefore, we investigate how selective dialog-frame retention affects long-horizon IIGN performance. We consider three memory variants: Decision keyframe, Spatially local keyframe, and Dialog keyframe, which compress the full dialog history into a small set of selected frames. Specifically, the Decision keyframe strategy retains dialog frames generated within the most recent $K_t$ decision steps, the Spatially local keyframe strategy preserves dialog frames whose recorded positions fall within a radius $r$ (in meters) of the current decision location, and the Dialog keyframe strategy keeps the most recently generated $K_f$ dialog frames. We conduct this experiment on VL-LN Bench, and the results are shown in Tab.~\ref{tab:memory_retention}.

\begin{table}[t]
\centering
\small
\setlength{\tabcolsep}{4pt}
\renewcommand{\arraystretch}{1.12}
\caption{\textbf{Ablation on dialog-frame retention strategies.}}
\label{tab:memory_retention}
\begin{tabular}{llcccc}
\toprule
Strategy & Setting & SR$\uparrow$ & SPL$\uparrow$ & NQ$\downarrow$ & Tokens $\downarrow$ \\
\midrule
\multirow{4}{*}{\makecell[l]{Decision\\keyframe}}
& 50  & 43.4 & 24.36 & 3.32 & 3204 \\
& 100 & 41.2 & 22.62 & 3.08 & 3270 \\
& 200 & 42.0 & 23.22 & 3.01 & 3381 \\
& 400 & 41.0 & 22.62 & 2.74 & 3443 \\
\midrule
\multirow{4}{*}{\makecell[l]{Spatially local\\keyframe}}
& 3  & 41.6 & 24.02 & 3.13 & 3191 \\
& 6  & 40.2 & 22.83 & 2.92 & 3324 \\
& 12 & 39.6 & 22.05 & 2.79 & 3447 \\
& 24 & 40.4 & 23.03 & 2.74 & 3470 \\
\midrule
\multirow{4}{*}{\makecell[l]{Dialog\\keyframe}}
& 2  & 41.2 & 23.88 & 3.17 & 3279 \\
& 4  & 41.0 & 22.66 & 2.71 & 3370 \\
& 8  & 39.8 & 22.81 & 2.69 & 3444 \\
& 16 & 38.4 & 21.90 & 2.63 & 3452 \\
\bottomrule
\end{tabular}

\begin{minipage}{0.45\linewidth}
\footnotesize
\textbf{Note.}
Tokens denotes the average number of input tokens.
\end{minipage}
\end{table}

The results reveal a consistent trade-off across the three dialog-frame retention strategies: retaining fewer dialog frames generally improves navigation performance but increases the number of clarification questions. For Decision keyframe, the shortest window of 50 decision steps achieves the best SR and SPL, reaching 43.4\% and 24.36\%, but increases NQ to 3.32. In contrast, expanding the window to 400 reduces NQ to 2.74, while lowering SR and SPL to 41.0\% and 22.62\%. Similar trends appear for Spatially local keyframe and Dialog keyframe: more compact dialog-frame memory reduces noisy grounding evidence and benefits decision-making, but may discard previously disambiguated instances, causing the agent to ask repeated clarification questions.

Among the three dialog-frame retention strategies, Decision keyframe achieves the best performance. Its 50-step setting obtains the best SR and SPL among all variants, outperforming the strongest settings of Spatially local keyframe and Dialog keyframe while using a comparable number of input tokens. This suggests that Decision keyframe better aligns with the temporal nature of interactive instance grounding: it preserves sufficient visual continuity for target tracking, whereas Spatially local keyframe may discard visible but distant instance evidence, and Dialog keyframe may suffer from unstable tracking in instance-dense scenes.

\subsection{Case Study}
\begin{figure}[t]
    \centering
    \includegraphics[width=\columnwidth]{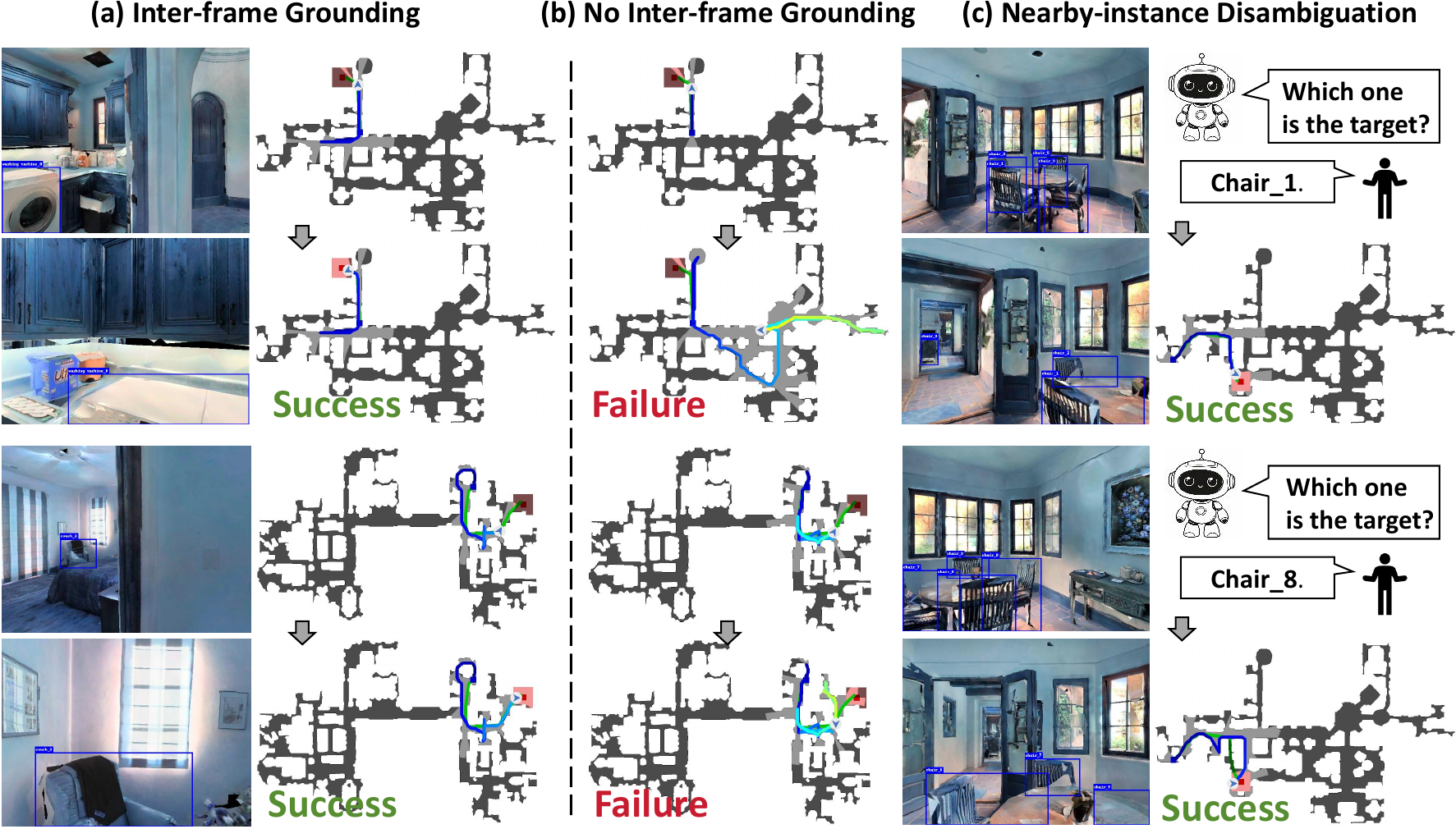}
    \caption{\textbf{Case studies on using intermediate observations and disambiguating similar targets a in distractor-rich scene.}
    }
    \label{fig:case_study}
\end{figure}

To better understand where GG-Nav provides the most substantial benefits, we present qualitative case studies in two representative scenarios: intermediate-frame grounding ~(Fig.~\ref{fig:case_study}(a)-(b)) and similar-instance disambiguation~(Fig.~\ref{fig:case_study}(c)). We first examine episodes in which the target instance appears only in intermediate frames between two decision steps, highlighting the importance of incorporating intermediate observations into the decision-making process. We then analyze target-dense scenes that contain multiple nearby and visually similar instances from the same category, demonstrating that GG-Nav can robustly disambiguate visually similar instances and maintain the correct referent over time. These two scenarios qualitatively demonstrate that GG-Nav addresses the two issues identified above by fully leveraging intermediate-frame evidence and accurately grounding the target instance despite distractors.

Intermediate observations are critical for decision making because the target may appear only briefly between adjacent decision steps. The paired examples in Fig.~\ref{fig:case_study}(a)-(b) compare episodes with and without intermediate observations. In Fig.~\ref{fig:case_study}(a), the first column shows the observations and grounding results when the target first appears and at the final step, while the second column and Fig.~\ref{fig:case_study}(b) show the corresponding exploration trajectories. Without intermediate-frame observations, briefly visible targets may be missed, causing the agent to treat the region as target-free and move away from the goal.

\begin{figure*}[htbp!]
    \centering
    \includegraphics[width=\textwidth]{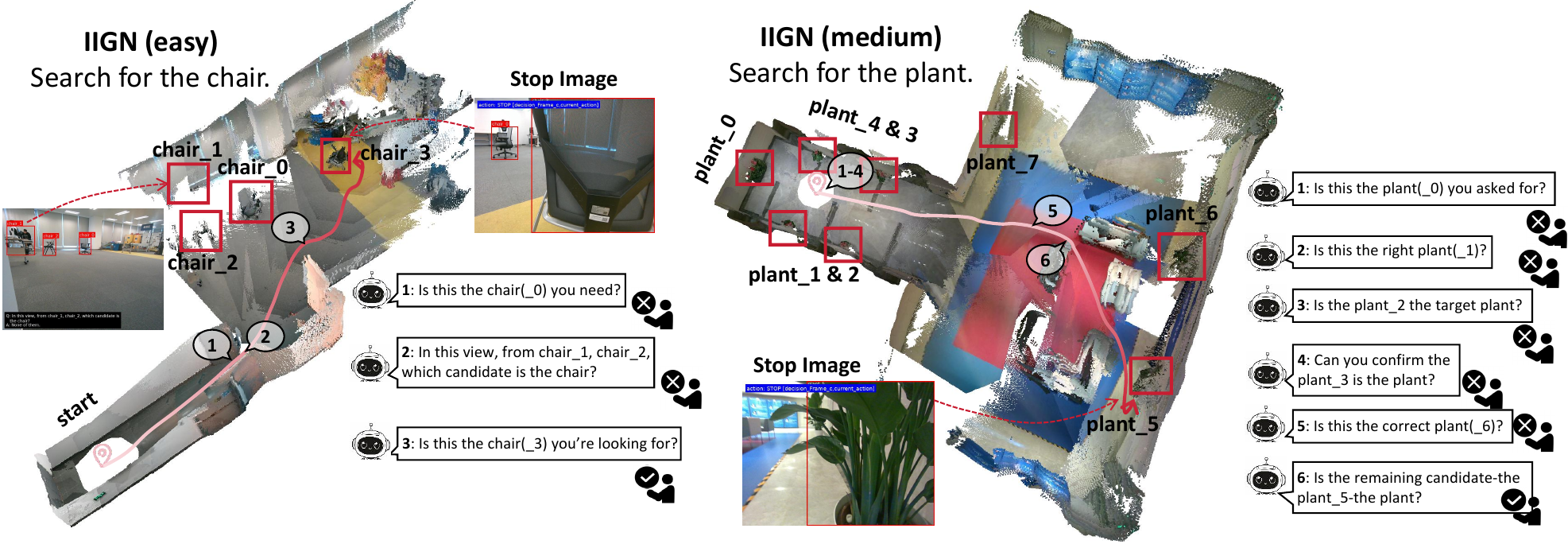}
    \caption{\textbf{Real-world navigation visualization.} The left side shows the IIGN (easy) scenario, where the agent is required to search for `chair\_3' among three distractors. The right side shows the IIGN (medium) scenario, where the robot needs to reach `plant\_5' among seven distractors. The robot-human dialogues are shown on the right, while selected observation frames and the final stopping frames are displayed next to the reconstructed scenes.}
    \label{fig:real_world}
\end{figure*}

GG-Nav can also follow human clarification and stop at the correct target in the presence of visually similar nearby distractors. As shown in Fig.~\ref{fig:case_study}(c), multiple chairs appear around the same table, making it challenging to identify and track the intended target during navigation. After the agent asks for clarification, the human specifies the target instance, such as \textit{Chair\_1} or \textit{Chair\_8}. In both cases, the agent successfully preserves the clarified referent, navigates to the selected chair, and stops near the correct target while facing it. This demonstrates that the proposed grounding mechanism can maintain instance-level consistency even in target-dense scenes with strong visual ambiguity.

\subsection{Real-world Experiment} 

\textbf{Experiment Setup.} We conduct real-world experiments on a Unitree Go2 quadruped robot. The robot is equipped with an Intel RealSense D455 RGB-D camera mounted at a height of 0.7 m and tilted downward by $15^\circ$. GG-Nav runs on a remote server with an RTX 4090 GPU. Given a goal-oriented instruction, the robot streams RGB-D observations to the server for inference. The model predicts either a navigation action or a clarification question. A navigation action is a discrete view adjustment or a pixel goal. Pixel goals are back-projected with RGB-D observations, transformed into world-frame waypoints using odometry, and tracked by an onboard MPC controller; discrete actions are executed by a PID-mode pose controller. When the model asks a clarification question, a human provides the answer through the server interface.

\begin{figure}[htbp!]
    \centering
    \includegraphics[width=0.618\columnwidth]{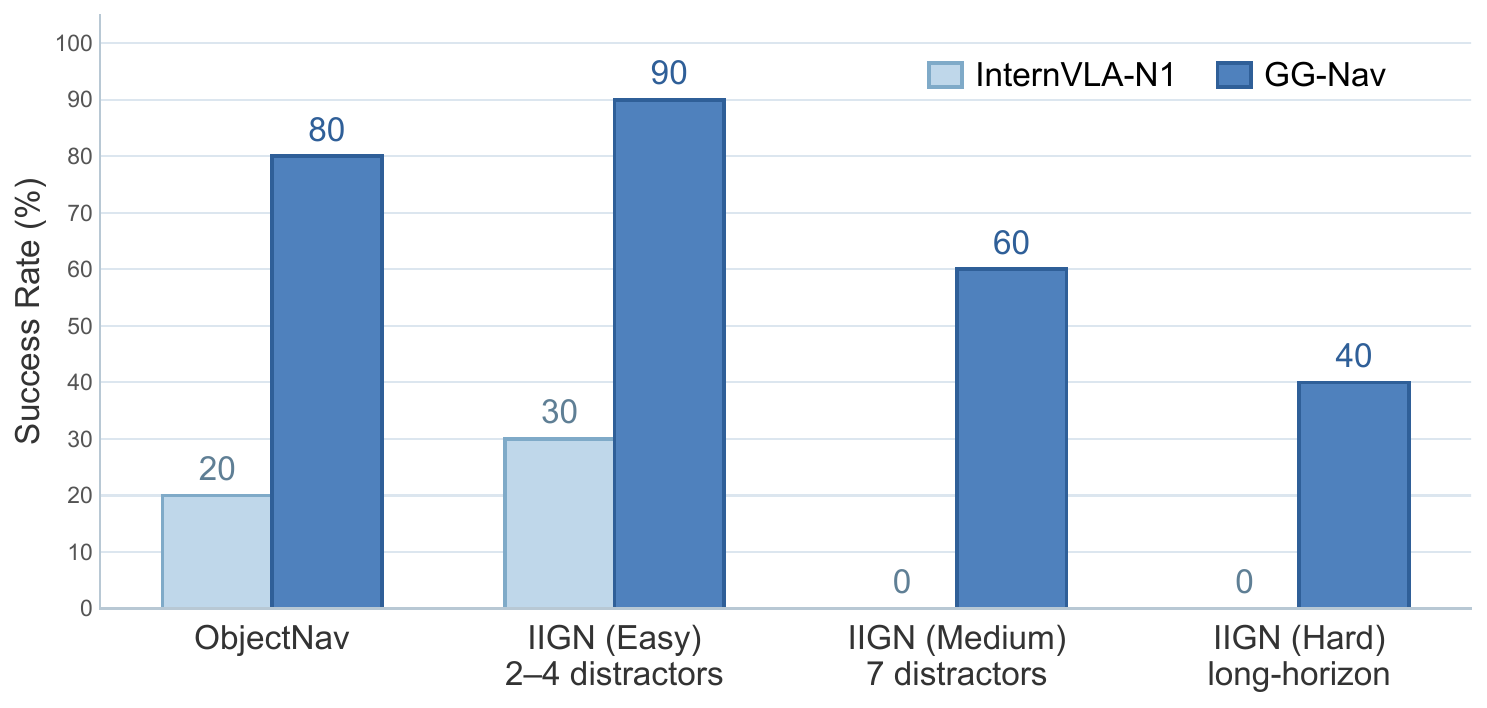}
    \caption{\textbf{Real-world navigation success rates.}}
    \label{fig:real_world_success_rate}
\end{figure}

\textbf{Quantitative Analysis.} To quantitatively evaluate the robustness and generalization of GG-Nav in real-world settings, we compare it with InternVLA-N1~\cite{internvla-n1} on ObjectNav and three IIGN settings with increasing ambiguity and navigation difficulty. Performance is measured by Success Rate (SR); an episode is successful only if the robot stops within $0.25$\,m of, and faces, the intended target instance. To better visualize the long-horizon interactive exploration process, we use VGGT-$\Omega$~\cite{wang2026vggtomega} to reconstruct the scene. This reconstruction is used only for visualization; GG-Nav itself does not require map construction during navigation. The ObjectNav and IIGN (hard) settings are shown in Fig.~\ref{fig:teaser}, while the IIGN (easy) and IIGN (medium) settings are shown in Fig.~\ref{fig:real_world}. Fig.~\ref{fig:real_world_success_rate} presents the quantitative comparison between GG-Nav and InternVLA-N1, showing that GG-Nav consistently outperforms InternVLA-N1 across ObjectNav and all IIGN settings. 

\textbf{Qualitative Analysis.} Detailed navigation processes of GG-Nav and InternVLA-N1 are provided in the supplementary video. We observe that InternVLA-N1 often fails to effectively rule out each potential target candidate and to stop within the $0.25$\,m success threshold. By converting intermediate frames into accumulated grounding evidence through Observation-to-Evidence, GG-Nav retains transient observations of potential target instances. As shown in Fig.~\ref{fig:real_world}, in the IIGN (easy) setting, the agent actively asks clarification questions about `chair\_1' and `chair\_2', even though each instance is visible in only a single frame during the entire process. In the IIGN (medium) setting, the agent starts in a dense cluster of visually similar plants, yet it can still accurately distinguish different instances and ask targeted clarification questions for disambiguation. Moreover, as shown in the stopping frames of both settings, GG-Nav can accurately stop near the target instances while facing them.

%% file: sections/conclusion.tex
\section{Conclusion}

We have presented GG-Nav, a navigation framework that learns a structured CoT mechanism to address insufficient observation utilization and imprecise target localization in goal-oriented navigation. Our results highlight that reliable navigation depends not only on the generation capability of vision-language models, but also on preserving transient visual evidence and maintaining consistent target grounding throughout navigation. Evaluations across ObjectNav and IIGN benchmarks demonstrate that GG-Nav consistently improves navigation and stopping reliability, while its deployment on a Unitree Go2 robot further validates its applicability to unseen real-world environments. These findings show that explicitly extracting goal-relevant visual information as structured evidence and reasoning over this evidence can improve the reliability of navigation decisions. By supporting the reliable execution of individual goal-oriented navigation subtasks, our proposed framework further lays a foundation for long-horizon navigation, whose tasks can often be decomposed into a sequence of such subtasks.

%% file: sections/appendix.tex
\section{Implementation Details}
\label{app:implementation}

We provide the training setup, data mixture, and inference configuration needed to
reproduce GG-Nav.

\subsection{Training Setup}

GG-Nav is initialized from Qwen3-VL-4B-Instruct. We optimize the vision encoder,
multimodal projector, and language-model backbone for 2.5 epochs using AdamW.
The learning rate is $2\times10^{-5}$ for the language-model backbone and
multimodal projector and $5\times10^{-6}$ for the vision encoder. We use a global
batch size of 128, zero weight decay, a warm-up ratio of 0.03, a cosine
learning-rate schedule, and a maximum gradient norm of 1. Training uses bfloat16,
gradient checkpointing, and DeepSpeed ZeRO-3, with a maximum sequence length of
8,192 tokens. Input RGB images are resized to $384\times384$. Following the visual-token
constraints of Qwen3-VL, the minimum and maximum input sizes are set to 3,136 and
313,600 pixels, respectively.

\subsection{Training Data Mixture}

The source corpus contains approximately 30K videos with grounding annotations
and 60K navigation trajectories, comprising 20K ObjectNav trajectories and 40K
IIGN trajectories. As described in Sec.~\ref{sec:data_collection}, each
source sequence yields multiple training instances. The resulting training set contains
1,000,000 temporal-grounding samples, 1,178,492 ObjectNav samples, and
3,543,549 IIGN samples. Thus, the 5,722,041-sample mixture comprises 17.5\%,
20.6\%, and 61.9\% of these three supervision types, respectively.

\subsection{Inference Configuration}
\label{app:inference}

\noindent\textbf{Visual context.}
At each decision step, we uniformly sample eight historical frames from the
start of the episode through the preceding decision step. We additionally retain
key frames containing grounded target-category instances, together with the
associated dialogue. All observations collected since the preceding decision are
passed to Observation-to-Evidence. Unless stated otherwise, the benchmark
experiments use the Decision-keyframe retention strategy with a window of 200
decision steps.

\noindent\textbf{Action space.}
The navigation action space comprises view adjustment, pixel-goal navigation, and
\texttt{STOP}. A view-adjustment action turns the agent left or right by an angle
in $\{30^\circ,60^\circ,90^\circ,120^\circ\}$. The model predicts a pixel goal as
a normalized horizontal--vertical coordinate on a $[0,1000]$ grid; this output is
then converted to an image coordinate $[w,h]$, where $0\leq w<W$ and
$0\leq h<H$. Depth and camera intrinsics are used to back-project this coordinate
to a local waypoint. Habitat's shortest-path follower executes the waypoint in
simulation, whereas the real robot uses the MPC/PID control stack described in
Appendix~\ref{app:realworld}. In IIGN, the model may instead produce a
natural-language clarification question. Before predicting a pixel goal, it emits
a downward-arrow token to request a downward-tilted observation; the corresponding
simulation action changes camera elevation by $30^\circ$.

\noindent\textbf{Decision timing.}
VLM inference is event-driven. After a pixel goal, the local controller executes a fixed number of low-level steps before
the next VLM call. After a view adjustment, the requested rotation is completed
before inference resumes. Following a clarification question, the model is
invoked again after the answer has been appended to the dialogue history. Consequently, high-level decisions occur at
action-dependent intervals rather than at a fixed frame rate.

\noindent\textbf{Decoding and failure handling.}
For benchmark evaluation, we generate at most 2,048 new tokens using stochastic
decoding with temperature $0.7$ and nucleus-sampling threshold $p=0.9$. We parse
the generated output into grounding records and an action. If parsing fails or
the action is invalid, the model is invoked again with the same decision context,
for at most three attempts in total. An episode is marked unsuccessful if all
three attempts fail.

\section{Structured Goal-Oriented Grounding CoT}
\label{app:cot_format}

\subsection{Grounding Output Format}

GG-Nav represents each frame-level grounding result as a JSON list. Every detected
target-category instance has the form
\texttt{\{"bbox\_2d":[$x_{\min}$,$y_{\min}$,$x_{\max}$,$y_{\max}$],
"label":"$c_k$"\}}, where $c$ is the instructed category and $k$ is a
model-assigned, temporally consistent track ID. Bounding-box coordinates are
normalized to a $[0,1000]$ scale and ordered as
$[x_{\min},y_{\min},x_{\max},y_{\max}]$. If no target-category instance is
detected in a frame, the model outputs an empty list, \texttt{[]}.

\subsection{Prompt and Output Examples}

\noindent\textbf{Observation-to-Evidence.}
The input consists of the compact visual history, the previous look-down frame,
and the recent observations collected before the current decision. The model
outputs one grounding list for each recent observation, in the same temporal
order, followed by a summary of newly observed or still-uncertain instances. For
example, its prompt is:

\begin{lstlisting}[basicstyle=\ttfamily\footnotesize,breaklines=true,
columns=fullflexible,keepspaces=true,showstringspaces=false]
You are an autonomous navigation assistant. Your task is to search for the tv. These are your historical observations: <image> x 8. Your previous look-down image is <image>. Now, first track object identities in the recent observations before the current decision: <image> x 3. These tracking images are only for maintaining consistent identities, matching previously seen instances, and summarizing newly discovered instances. Do not decide the final action yet.
\end{lstlisting}

For three recent observations of the same television, the corresponding output is:

\begin{lstlisting}[basicstyle=\ttfamily\footnotesize,breaklines=true,
columns=fullflexible,keepspaces=true,showstringspaces=false]
<think>
```json
[{"bbox_2d":[814,504,1000,652],"label":"tv_0"}]
```
```json
[{"bbox_2d":[471,502,614,625],"label":"tv_0"}]
```
```json
[{"bbox_2d":[110,502,287,643],"label":"tv_0"}]
```
I found 1 uncertain tv(s) in the recent tracking images.
Uncertain instances: tv_0.
</think>
\end{lstlisting}

Although the bounding box changes across viewpoints, the identifier
\texttt{tv\_0} remains consistent. This stage only accumulates grounding evidence
and does not predict an action.

\noindent\textbf{Evidence-to-Decision.}
The current observation is appended after the Observation-to-Evidence output. The
model grounds the current frame and then generates a view-adjustment action, pixel
goal, \texttt{STOP}, or clarification question. When asking about accumulated
evidence, the prefix \texttt{Frame\_k} binds a question to the $k$-th intermediate
frame, whereas \texttt{Frame\_c} denotes the current frame. An abbreviated prompt
is:

\begin{lstlisting}[basicstyle=\ttfamily\footnotesize,breaklines=true,
columns=fullflexible,keepspaces=true,showstringspaces=false]
Your current image is <image>. First ground the target-category instances in this image, and then decide the next action from all available context. You may move or talk. Bind a question about an intermediate image to "Frame_k" and a question about the current image to "Frame_c". To output a waypoint, first request a downward-tilted observation and then predict its image coordinate. If the waypoint is outside the current view, use TURN LEFT or TURN RIGHT. Output STOP only after the task has been completed.
\end{lstlisting}

\noindent\emph{Dialogue-action example.}
When the target is absent from the current frame but a candidate exists in the
accumulated evidence, the model can ask a frame-grounded question:

\begin{lstlisting}[basicstyle=\ttfamily\footnotesize,breaklines=true,
columns=fullflexible,keepspaces=true,showstringspaces=false]
<think>
I did not detect any tv in the current image.
</think>
<talk>Frame_1: Is this one the tv?</talk>
\end{lstlisting}

Here, the current-frame grounding is empty, but the previously accumulated
\texttt{tv\_0} evidence remains available. The model therefore asks about the
candidate in \texttt{Frame\_1} rather than discarding it or generating an
ungrounded question.

\noindent\emph{Navigation-action example.}
When the target is grounded in the current observation, the model first requests
an observation tilted downward by $30^\circ$. It then predicts a pixel goal on
this look-down frame:

\begin{lstlisting}[basicstyle=\ttfamily\footnotesize,breaklines=true,
columns=fullflexible,keepspaces=true,showstringspaces=false,mathescape=true]
Assistant:
<think>Target tv found. Navigate to it.</think><move>$\downarrow$

Input:
<image>  (30-degree look-down frame)

Assistant:
<move>371 550
\end{lstlisting}

Here, \texttt{371 550} is the normalized horizontal--vertical coordinate on the
$[0,1000]$ grid. It is converted to the resolution of the look-down image and
back-projected with depth and camera intrinsics to obtain a local waypoint for
execution.

\subsection{VLM-based Instance Association}

The VLM predicts visible target-category instances and their temporally consistent
track IDs from historical grounding records, retained key frames, and recent
observations. It reuses an existing ID when the context supports an identity match
and otherwise creates a new ID. We do not use an external object tracker,
geometric matcher, or post-hoc identity-association module. Dialogue answers are
stored with the corresponding grounded frame and track ID for subsequent
decisions.

\section{Training Data Construction}
\label{app:data}

\subsection{Videos with Grounding Annotations}

We construct Observation-to-Evidence training samples from continuous egocentric
videos in the Aria Synthetic Environments Dataset. For each video, we sample an
object category present in the scene, select $K$ context frames containing at least
one instance of that category, and extract a contiguous clip of $L$ frames, where
$K\in\{1,\ldots,8\}$ and $L\in\{1,2,3,7\}$. We retain an instance only when its
visible segmentation mask contains at least 200 pixels and convert the valid mask
to its enclosing 2D bounding box. The context annotations condition the model,
whereas the clip boxes, instance IDs, and evidence summary form the supervision.
The resulting instances constitute the temporal-grounding portion of the training
mixture summarized in Appendix~\ref{app:implementation}.

\subsection{Navigation Trajectories}

We generate IIGN episodes from all training scenes of VL-LN Bench and ObjectNav
episodes from all training scenes of HM3D ObjectNav. The simulated agent uses
$640\times480$ RGB-D observations with a $79^\circ$ horizontal field of view. The
camera is mounted at 1.25\,m, and the depth range is 0--10\,m. The agent radius is
0.18\,m, with a 0.25\,m forward step and a $30^\circ$ turn angle.

During data collection, a frontier-based exploration policy repeatedly selects waypoints along the boundary of the explored free space, while an A$^\ast$ planner converts each waypoint into a collision-free action sequence. We record the observations, agent poses, actions, bounding boxes, and simulator instance IDs of all visible target-category objects. We retain only trajectories that satisfy two criteria: the target is visible in the final observation, and the trajectory contains no redundant exploration. The retained trajectories are then used to supervise both Observation-to-Evidence and Evidence-to-Decision learning.

\subsection{Interaction Supervision}

During navigation, we maintain a set of observed instances belonging to the target category and track whether each instance has been confirmed or ruled out through previous interactions. Whenever the agent observes one or more previously unseen candidate instances, we generate a question based on the number of newly observed candidates and the elimination history of previously observed ones. Representative question templates include:
\begin{itemize}
    \item \textit{Single candidate:} ``Is this the target
    \{category\}?''
    \item \textit{Multiple unseen candidates:} ``Which of these
    \{category\} instances is the target?''
    \item \textit{Partially eliminated candidates:} ``After ruling out \{category\} instances \{eliminated\_instance\_ids\}, which of the remaining \{category\} instances is the target?'' When only one candidate remains, we instead ask, ``Is \{remaining\_instance\_id\} the target after ruling out \{eliminated\_instance\_ids\}?''
\end{itemize}
The automated oracle resolves each referenced candidate using Habitat's
ground-truth instance-segmentation mask and compares its simulator instance ID
with the annotated target ID. It answers a binary question with \texttt{Yes} or
\texttt{No}. For a multiple-candidate question, it returns the target's referenced
ID when present and \texttt{None of them} otherwise. Ground-truth segmentation is
used only to generate the answer; the answer exposes no privileged mask,
coordinate, pose, distance, or directional information. The question--answer pair
is then appended to the dialogue history.

\begin{figure*}[t]
    \centering
    \begin{subfigure}[t]{0.32\textwidth}
        \centering
        \includegraphics[width=\linewidth]{%
            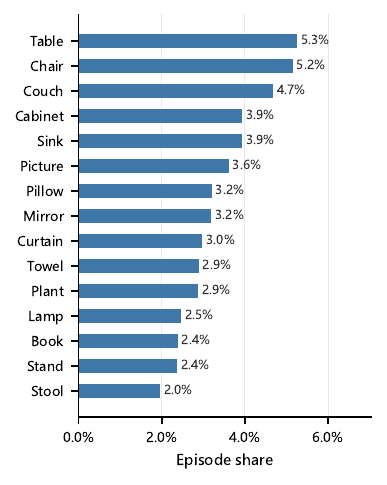}
        \caption{The 15 most frequent target categories.}
        \label{fig:iign-category-distribution}
    \end{subfigure}
    \hfill
    \begin{subfigure}[t]{0.32\textwidth}
        \centering
        \includegraphics[width=\linewidth]{%
            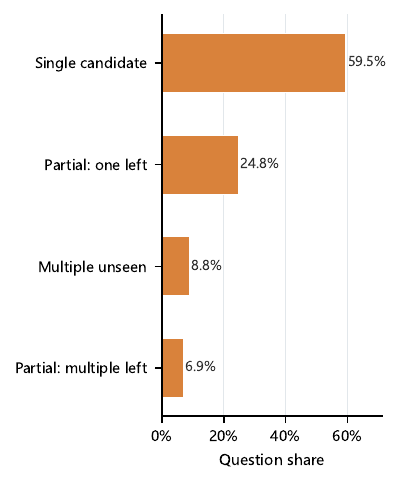}
        \caption{Candidate configuration.}
        \label{fig:iign-question-type-distribution}
    \end{subfigure}
    \hfill
    \begin{subfigure}[t]{0.32\textwidth}
        \centering
        \includegraphics[width=\linewidth]{%
            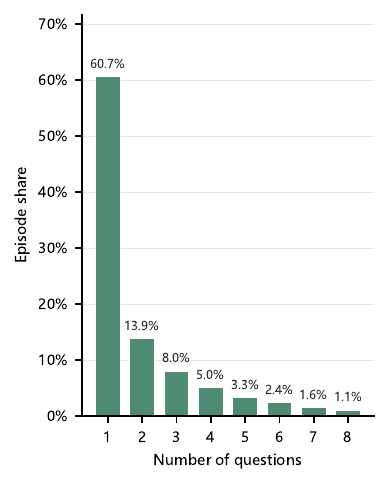}
        \caption{Clarification turns per episode.}
        \label{fig:iign-dialogue-turn-distribution}
    \end{subfigure}
    \caption{\textbf{Statistics of IIGN interaction supervision.}
    \textbf{(a)} The 15 most frequent target categories, ranked by episode
    count; bar annotations report their shares of all 50,855 episodes.
    \textbf{(b)} The distribution of 118,309 clarification questions over
    candidate configurations. ``Partial'' denotes that some candidates have
    already been eliminated, while ``one/multiple left'' indicates whether the
    remaining uncertain set contains one or multiple candidates (two in this
    corpus). \textbf{(c)} The number of clarification
    turns per episode. For readability, only bins whose individual shares are
    at least 1\% are shown; the omitted bins contain 1,961 episodes (3.86\%).}
    \label{fig:iign-supervision-statistics}
\end{figure*}

Figure~\ref{fig:iign-supervision-statistics} summarizes the IIGN
interaction-supervision pool before final trajectory selection. It contains
50,855 candidate episodes and 118,309 question--answer pairs; approximately 40K
IIGN trajectories from this pool are included in the final training mixture.

The 15 most frequent categories account for 50.97\% of episodes, while the
remaining 147 categories form a long tail. The most frequent individual category
accounts for only 5.26\%, indicating that the corpus is not dominated by a single
object class. Candidate configurations are more concentrated:
single-candidate questions constitute 59.47\% of all questions, and cases with one
candidate remaining after partial elimination contribute another 24.77\%. Thus,
84.24\% of interactions are binary confirmations. The remaining 15.76\% involve
multiple candidates or rejection of all visible candidates: 8.84\% contain
multiple previously unseen candidates, and 6.91\% retain multiple candidates
after partial elimination.

Dialogue length is similarly skewed toward concise interactions: 60.73\% of
episodes contain one clarification turn and 13.90\% contain two, while the mean is
2.33 turns per episode. The retained multi-turn tail nevertheless exposes the
model to repeated clarification and evidence updates rather than only isolated
single-turn confirmations.

\section{Evaluation Protocol}
\label{app:evaluation}

\subsection{Habitat Agent Configuration}
\label{app:habitat_config}

We use the same simulated embodiment and action space for all three benchmarks:
HM3D ObjectNav, VL-LN, and CoIN. In Habitat, the robot is modeled as a cylindrical
agent with a height of 1.25\,m and a radius of 0.18\,m. It is equipped with
co-located RGB and depth sensors mounted at 1.25\,m. Both sensors produce
$640\times480$ observations with a $79^\circ$ horizontal field of view, and the
depth sensor has a measurement range of 0--10\,m. GPS and compass sensors are also
enabled in the task configuration.

The discrete action space allows the agent to stop, move forward, turn left or
right, and look up or down. A forward action moves the agent by 0.25\,m, a turning
action changes its heading by $30^\circ$, and a look action changes the camera
elevation by $30^\circ$. Sliding along obstacles is enabled, and each episode is
limited to 500 simulator steps. An episode is successful when the agent issues
\texttt{STOP} at a robot-to-target Euclidean distance of at most 0.25\,m. Any
instance of the instructed category is valid in ObjectNav; only the annotated
target instance is valid in IIGN.

\subsection{Baseline Selection and Adaptation}

We select five representative baselines that cover the two dominant paradigms in
goal-oriented navigation: modular systems and end-to-end learned policies. The
modular baselines isolate the effects of geometric exploration, semantic guidance,
and explicit interaction, whereas the end-to-end baselines provide direct
comparisons with learned goal-oriented policies and navigation foundation models.
All results are reproduced under the common Habitat configuration described in
Appendix~\ref{app:habitat_config}, rather than quoted from the original papers.

\textbf{Modular methods.} FBE is a geometry-only frontier-based exploration
baseline derived from the VLFM implementation~\cite{yokoyama2024vlfm}. We remove
VLFM's language-grounded value map while retaining its mapping, target-detection,
and local-navigation components. Frontier selection is therefore based only on
geometric exploration cues, providing a language-agnostic reference for measuring
the benefit of semantic guidance. VLFM~\cite{yokoyama2024vlfm} is a representative
zero-shot modular ObjectNav method. It builds an occupancy map from depth
observations, extracts frontiers at the boundary between explored and unexplored
space, and scores the observed regions against the target category with a
pre-trained vision-language model. The resulting language-grounded value map
guides the agent toward the most promising frontier, while a pre-trained detector
and local policy handle target localization and motion.

AIUTA~\cite{CoIN} complements these exploration-oriented baselines with an explicit
interaction mechanism. It is a training-free modular method that operates
independently of the underlying navigation policy. Its Self-Questioner uses
LLM--VLM self-dialogue and uncertainty estimation to refine the description of a
candidate instance, after which an Interaction Trigger decides whether to ask the
user, continue navigation, or stop. We include AIUTA as a representative modular
baseline for interaction-driven instance disambiguation in IIGN.

\textbf{End-to-end methods.} MTU3D~\cite{zhu2025move} is designed directly for
goal-oriented embodied navigation. It constructs an online query-based spatial
memory from RGB-D observations and jointly represents grounded objects and
unexplored frontiers, thereby coupling visual grounding, frontier selection, and
trajectory prediction in a unified learned framework. Because its original model
already supports category-, language-, and image-specified navigation goals, we
evaluate the released pretrained weights directly without task-specific
fine-tuning.

InternVLA-N1~\cite{internvla-n1} is an open navigation foundation model and serves
as the strongest learned baseline in our comparison. It adopts a dual-system
architecture in which a vision-language System~2 predicts mid-term pixel goals or
latent plans and a diffusion-based System~1 executes short-horizon motion. The
released model, however, is primarily trained for vision-and-language navigation
and does not natively cover category-level ObjectNav or interactive instance
disambiguation. We therefore initialize its System~2 from the official weights and
fine-tune only this high-level component with our ObjectNav and IIGN navigation
data, aligning its goal inputs and outputs with the two task formulations. In
simulation, we do not use InternVLA-N1's learned System~1; instead, its predicted
pixel goals are executed by Habitat's shortest-path follower.

For evaluation, every method uses the same sensor configuration, robot embodiment,
episode budget, and success threshold, while consuming the subset of RGB-D inputs
required by its native architecture. For the direct comparison between
InternVLA-N1 and our method, InternVLA-N1's System~2 and GG-Nav serve as the
respective high-level decision modules, while both share the same Habitat
shortest-path follower as System~1. This controlled low-level execution isolates
differences in high-level goal-oriented reasoning and decision making. On IIGN,
the non-interactive baselines receive the same underspecified initial instruction
but cannot ask clarification questions; success is awarded only for reaching the
annotated target instance rather than an arbitrary instance of the same category.
This setting quantifies how far standard ObjectNav policies can progress without
interaction or explicit instance disambiguation. AIUTA and InternVLA-N1 are
allowed to query the same interaction oracle, thereby holding oracle access fixed
across the interactive methods.

\subsection{Stage-wise Failure Attribution}
\label{app:failure_rules}

Each failed episode is assigned to one stage: exploration, grounding, tracking,
reaching, or stopping. Attribution is fully automated from the simulator's
ground-truth instance-segmentation masks, the model's grounding/tracking records,
and the final robot pose; no manual labels are used. Let
$d_{\mathrm{final}}$ denote the straight-line robot-to-target distance at
termination. We apply the following rules in order, assigning each failed episode
to the first rule that it satisfies:
\begin{enumerate}
    \item \textbf{Exploration:} the target's ground-truth instance ID is absent
    from every instance-segmentation observation; equivalently, the target never
    enters the camera view.
    \item \textbf{Grounding:} the target appears in at least one observation, but
    none of the model's grounding records is associated with its ground-truth
    instance ID.
    \item \textbf{Tracking:} the target is correctly grounded at least once, but
    its established track ID is not recovered when the target next appears, or
    that ID is subsequently assigned to a different instance.
    \item \textbf{Reaching:} the correct target association is established and
    retained, but $d_{\mathrm{final}}\geq1$\,m; the target neighborhood is defined
    as the set of robot poses less than $1$\,m from the target.
    \item \textbf{Stopping:} the final stopping position is inside the target
    neighborhood but outside the success region, i.e.,
    $0.25<d_{\mathrm{final}}<1$\,m.
\end{enumerate}

\section{Real-world Experiment Details}
\label{app:realworld}

\subsection{System Architecture and Execution}

Our platform is a Unitree Go2 equipped with an Intel RealSense D455 RGB-D camera
mounted at $0.7$\,m and tilted downward by $15^\circ$. A ROS~2 client
time-synchronizes the $640\times480$ RGB and aligned-depth streams with a 0.1\,s
tolerance and associates each observation with the nearest odometry sample. Each
RGB-D observation is sent over HTTP to GG-Nav on a remote RTX~4090 server, while
low-level control runs onboard the robot.

For each high-level decision, GG-Nav uses eight uniformly sampled historical
frames and all dialogue frames. It also samples at most seven
observations collected since the preceding decision, first grounding and tracking
instances in these intermediate frames and then deciding from this evidence and
the current frame. The model emits a discrete action, a pixel goal, or a
clarification question. Movement commands are returned to the robot client,
whereas an IIGN question is handled on the server: it explicitly references the
current or an intermediate frame, the corresponding image is displayed with its
grounding boxes and instance IDs, and the human answer is attached to that frame
in the subsequent history. Because the robot uses a fixed downward-tilted camera,
a look-down request triggers a second inference stage on the fixed camera stream
rather than a physical camera motion.

For a pixel goal, the server denormalizes the predicted coordinate to the
$640\times480$ image, back-projects it using aligned depth and camera intrinsics,
compensates for camera pitch, and limits the resulting robot-frame waypoint to
$1$\,m. The client transforms this local waypoint into the world frame using the
time-aligned odometry and tracks it with a receding-horizon MPC controller
(20 steps at $0.1$\,s, with $|v|\leq0.3$\,m/s and
$|w|\leq0.3$\,rad/s). Discrete actions are converted into pose increments from
the current odometry---$0.25$\,m forward,
$30^\circ$ left/right, or stop---and executed by a PID-mode pose controller with
maximum linear and angular speeds of $0.6$\,m/s and $0.5$\,rad/s, respectively.
After a turn, the next server query is delayed until the yaw error is below
$5^\circ$ and angular speed below $0.08$\,rad/s, or until a $10$\,s timeout.
Both controllers ultimately publish $(v,0,w)$ commands through the Unitree Sport
API.

\subsection{Scenarios and Evaluation}

We evaluate four real-world settings: ObjectNav and three IIGN difficulty levels.
The IIGN levels are defined by the navigation distance and the number of
same-category candidate instances requiring disambiguation:
\begin{itemize}
    \item \textbf{Easy:} navigation distance $<50$\,m and fewer than five
    candidate instances.
    \item \textbf{Medium:} navigation distance $<50$\,m and at least five
    candidate instances.
    \item \textbf{Hard:} navigation distance $>50$\,m and at least five
    candidate instances.
\end{itemize}

An episode is successful only when the robot issues \texttt{STOP} within
$0.25$\,m of the intended target instance while facing that instance. The
final robot-to-target distance is measured manually in the laboratory, and the
requirement that the robot face the target is verified by visual inspection at
termination.

\section{Limitations and Future Work}
\label{app:limitations}

First, the current interaction policy supports only target-specific clarification
questions about observed candidate instances. Real-world interaction permits a
broader and potentially more efficient set of information-seeking behaviors. For
example, an agent could improve exploration by asking for route guidance, such as
``How can I get to the target instance?'', or improve disambiguation through a
compositional question that combines room context and nearby attributes, such as
``Is the target sofa in the living room and next to a red decoration?'' Future
work will extend GG-Nav from candidate verification to open-ended question
planning, allowing it to decide when to request navigational guidance and which
combination of semantic and spatial cues would most effectively reduce target
uncertainty.

Second, the current target vocabulary and object scale are constrained by the
Habitat-based task setup and its available object annotations. Consequently, our
experiments primarily cover relatively large, common household objects, such as
beds, tables, and mirrors, and we do not yet claim reliable support for small
everyday objects such as phones or forks. Extending GG-Nav to these targets will
require simulation and real-world training data with substantially greater
coverage of small objects, finer-grained instance annotations, and
higher-resolution or active close-range perception for robust grounding and
tracking.